\documentclass{bmvc2k}
\usepackage{xcolor} 
\hypersetup{pdfpagelayout=SinglePage}

\title{AD-GS: Anatomical Density-guided 3D Gaussian Splatting for Sparse-View Dental CBCT Reconstruction}

\addauthor{Zonglin Li\textsuperscript{*}*}{zonglinli@bupt.edu.cn}{1}
\addauthor{Kexian Xie\textsuperscript{*}*}{kexianxie@bjmu.edu.cn}{2}
\addauthor{Feiran Li}{frl1994@hotmail.com}{3}
\addauthor{Chu Zhou}{zhou_chu@hotmail.com}{4}
\addauthor{Yixing Huang}{huangyx@pku.edu.cn}{5}
\addauthor{Ruiyang Chen}{chenruiyang@bupt.edu.cn}{1}
\addauthor{Zhanyu Ma}{mazhanyu@bupt.edu.cn}{1}
\addauthor{Heng Guo\textsuperscript{\dag}}{guoheng@bupt.edu.cn}{1}

\addinstitution{
 Beijing University of Posts and Telecommunications\\
 Beijing, China
}
\addinstitution{
 Department of Stomatology, Peking University Third Hospital\\
 Beijing, China
}
\addinstitution{
 Independent Researcher
}
\addinstitution{
 National Institute of Informatics\\
 Tokyo, Japan
}
\addinstitution{
 Institute of Medical Technology, Peking University\\
 Beijing, China
}

\runninghead{Li, Xie, et al.}{AD-GS}

\usepackage{amsfonts}
\usepackage{amssymb}
\usepackage{bm}
\usepackage{booktabs}
\usepackage{multirow}
\usepackage{makecell}
\usepackage{colortbl}
\usepackage{wrapfig}
\usepackage{threeparttable}
\usepackage{adjustbox}
\usepackage{microtype}
\usepackage{nicefrac}
\usepackage{pdfpages}

\newcommand{\Tref}[1]{Table~\ref{#1}}

\newcommand{\fref}[1]{Fig.~\ref{#1}}
\newcommand{\Fref}[1]{Figure~\ref{#1}}

\usepackage{xcolor}
\newcounter{todos}
\AtEndDocument{\ifnum\value{todos}>0 \PackageWarning{TODOS}{There are \arabic{todos} todos left in this paper! Fix them before submitting the paper!} \fi}

\makeatletter
\DeclareRobustCommand\onedot{\futurelet\@let@token\@onedot}
\def\@onedot{\ifx\@let@token.\else.\null\fi\xspace}
\def\eg{\emph{e.g}\onedot} 
 \def\Ie{\emph{I.e}\onedot}

\makeatother
\usepackage[utf8]{inputenc} 
\usepackage[T1]{fontenc}    
\usepackage{url}            
\usepackage{booktabs}       
\usepackage{amsfonts}       
\usepackage{nicefrac}       
\usepackage{xcolor}         
\usepackage{graphicx}
\usepackage{amsmath}
\usepackage{amssymb}
\usepackage{bm}
\usepackage{makecell}
\usepackage{multirow}
\usepackage{xspace}
\usepackage{colortbl}
\usepackage{wrapfig}
\usepackage{lipsum} 
\usepackage{threeparttable}

\definecolor{MyDarkRed}{rgb}{0.46, 0.16, 0.16}
\definecolor{MyDarkBlue}{rgb}{0.16, 0.16, 0.66}
\definecolor{MyPink}{rgb}{1.0, 0.702, 0.729} 
\definecolor{MyPeach}{rgb}{1.0, 0.875, 0.729} 
\definecolor{MyLightYellow}{rgb}{1.0, 1.0, 0.729} 
\definecolor{MyLightGreen}{rgb}{0.729, 1.0, 0.788} 
\definecolor{MyLightBlue}{rgb}{0.729, 0.882, 1.0} 

\newcommand{\ours}{AD-GS\xspace}

\newcommand{\rrgs}{$R^2$-Gaussian~\cite{R2}\xspace}

\newcommand{\fdk}{FDK~\cite{FDK}\xspace}
\newcommand{\sart}{SART~\cite{SART}\xspace}
\newcommand{\asdpocs}{ASD-POCS~\cite{ASD-POCS}\xspace}

\newcommand{\naf}{NAF~\cite{zha2022naf}\xspace}

\newcommand{\saxnerf}{SAX-NeRF~\cite{SAX-NeRF}\xspace}

\newcommand{\toothseg}{ToothSeg~\cite{toothseg}\xspace}

\begin{document}

\maketitle

\renewcommand{\thefootnote}{}
\footnotetext[1]{*Equal contribution.\ \dag{} Corresponding author.}
\renewcommand{\thefootnote}{\arabic{footnote}}

\begin{abstract}

Sparse-view Cone-Beam Computed Tomography (CBCT) reconstruction is essential for reducing radiation dose and scan effort, but remains underconstrained due to limited projection information. Recent 3D Gaussian Splatting (3DGS)-based CBCT reconstruction methods rely on gradient-driven densification, which becomes unstable under sparse acquisition and often leads to over-smoothed structures. We propose AD-GS, an Anatomical Density-guided 3D Gaussian Splatting framework that formulates Gaussian densification as an attenuation-aware Markov Chain Monte Carlo (MCMC) sampling process. By interpreting Gaussians as samples from an underlying anatomical attenuation distribution, AD-GS adaptively concentrates representation capacity in high-attenuation regions, enabling sharper structural boundaries under sparse projections. To further constrain reconstruction, we incorporate a semantic regularization based on zero-shot features from Segment Anything in Medical Images, enforcing multi-view consistency through semantic priors. Experiments on the ToothFairy3 dental CBCT dataset demonstrate that AD-GS achieves superior structural fidelity and boundary preservation compared with existing 3DGS-based approaches under sparse-view settings within the dental and maxillofacial setting. Code is available at \url{https://github.com/yulinjiaxue/AD-GS-BMVC2026}.

\end{abstract}
\section{Introduction}
\label{sec:intro}

Cone-beam Computed Tomography (CBCT) is widely used in dental and maxillofacial diagnostics, enabling high-resolution 3D reconstruction of hard tissues such as teeth, enamel, cortical bone, and alveolar bone \cite{nasseh2018cone,CBCT}. To reduce radiation exposure under the As Low As Reasonably Achievable (ALARA) principle \cite{ALARA}, sparse-view acquisition is often preferred in clinical practice because it reduces the number of X-ray projections and shortens the scan procedure. This low-dose setting is especially attractive for repeated examinations and dental workflows, where radiation burden should be carefully controlled. In this work, we focus on this dental and maxillofacial CBCT setting.

However, sparse-view CBCT reconstruction constitutes a severely ill-posed inverse problem. Limited X-ray projections provide only partial angular constraints, making them insufficient to uniquely determine the underlying 3D attenuation field. Many different attenuation volumes may explain the same sparse measurements, and this ambiguity becomes especially severe around high-attenuation dental hard tissues (e.g., enamel and cortical bone), where small geometric errors can produce strong projection differences. Consequently, classical analytical reconstruction methods such as \fdk, while computationally efficient and widely used, tend to suffer from amplified noise, streak artifacts, and physical artifacts under sparse-view sampling conditions~\cite{R2,chen2026radioactive}. These artifacts can blur clinically important boundaries and degrade the reliability of subsequent structural measurements.

To mitigate the inherent ill-posedness associated with sparse-view CBCT reconstruction, a significant body of research has gravitated towards learning-based paradigms~\cite{Difgs}. These approaches typically leverage high-capacity neural architectures, such as Deep Convolutional Neural Networks (CNNs) and Vision Transformers (ViTs), to establish direct or indirect non-linear mappings from the under-sampled 2D projection domain to the high-dimensional 3D volumetric space. By supervising the network on large-scale clinical datasets, these models effectively learn sophisticated structural priors, enabling them to hallucinate missing sinogram information, suppress streaking artifacts, or refine noisy reconstructions in the image domain.

However, a fundamental challenge persists regarding the distributional shift, or domain gap, between curated training datasets and real-world clinical acquisitions. Beyond this generalization hurdle, the practical implementation of these supervised methods is further hindered by the immense difficulty in acquiring large-scale, high-quality real-world datasets. In the clinical domain, constructing a comprehensive and diverse corpus of paired sparse-view projections and high-fidelity 3D ground truths is often prohibitive. 

NeRF-based neural representations (\eg, \naf and \saxnerf) offer another direction by modeling the attenuation field as a continuous coordinate-based function optimized through differentiable volumetric rendering. These methods reduce the dependence on paired training data and can directly fit a specific scan from its projections. Nevertheless, dense ray sampling and repeated neural network evaluation incur substantial computational overhead, making them less practical for time-sensitive clinical workflows. In contrast, 3D Gaussian Splatting (3DGS) represents the scene using explicit anisotropic Gaussian primitives, which provides efficient rendering and a compact continuous representation. Recent methods such as \rrgs and X-Field~\cite{wangx} extend 3DGS to CBCT reconstruction by incorporating the Beer-Lambert law~\cite{beer1,beer2} to correct volume rendering bias. This makes 3DGS a promising foundation for sparse-view CBCT.

\begin{figure}[t]
    \centering
    \includegraphics[width=1.0\textwidth]{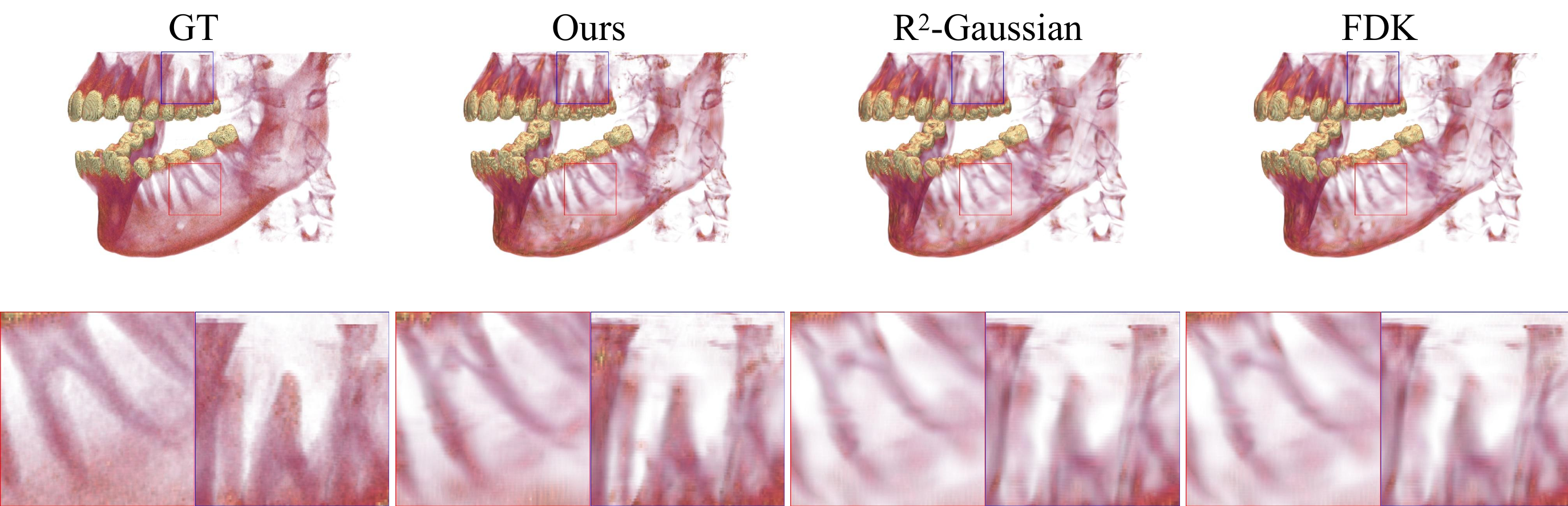} 
    \caption{AD-GS achieves superior fidelity and sharper details than both the 3DGS-based method \rrgs and classical analytical method \fdk in sparse~(30 views) CBCT reconstruction.}
    \label{fig:teaser}
\end{figure}


Nevertheless, existing 3DGS-based CBCT methods~\cite{R2,wangx} exhibit a shared limitation: they rely on gradient-driven ``clone-and-split'' densification that assumes large rendering gradients reliably indicate regions requiring refinement. This assumption is reasonable when observations are dense enough to provide stable multi-view constraints, but it becomes fragile under sparse-view settings. Each projection only constrains a limited set of rays through the volume, and the same local anatomical structure may be weakly observed or ambiguously explained from different angles. As a result, gradients from different views become inconsistent and may cancel each other out, leading to gradient collision~\cite{AbsGS}. Once this occurs, Gaussian primitives are either insufficiently allocated to high-frequency hard-tissue boundaries or are propagated toward suboptimal regions. The final reconstruction therefore tends to preserve global intensity trends while losing fine anatomical details that are critical for dental analysis. \Fref{fig:teaser} illustrates this effect on \rrgs, where bony structures appear blurry and smoothed compared to our method.

To address this limitation, we propose \emph{AD-GS}, an Anatomical Density-Guided 3D Gaussian Splatting framework for sparse-view CBCT reconstruction. Instead of triggering densification based only on local gradient magnitude, we reconsider Gaussian placement from the perspective of attenuation distribution. In CBCT, high-density structures such as enamel and cortical bone exhibit sharp attenuation transitions while occupying only a small portion of the volume. These regions are precisely where clinically meaningful geometry is concentrated: tooth crowns, roots, cortical boundaries, and narrow gaps between adjacent teeth all require sufficient representational capacity. Under sparse-view conditions, effectively allocating Gaussians to these high-attenuation regions is therefore crucial for preserving structural fidelity. This motivates us to use anatomical density as an explicit guide for Gaussian proliferation, rather than relying solely on unstable sparse-view gradients.

AD-GS improves sparse-view CBCT reconstruction through two complementary designs: 
\textbf{(i) attenuation-guided MCMC sampling} that reallocate Gaussian primitives 
toward structurally significant regions, and \textbf{(ii) semantic regularization} 
that injects foundation-model anatomical priors beyond projection supervision.

\paragraph{Attenuation-guided MCMC sampling.}
Specifically, AD-GS reinterprets the Gaussian set as samples from an underlying 
attenuation distribution and optimizes them within a Markov Chain Monte Carlo (MCMC) 
framework. On top of projection-consistency supervision, structured stochastic 
perturbations enable controlled exploration in high-attenuation regions. The sampling 
probability is modulated by the estimated attenuation magnitude, encouraging Gaussians 
to concentrate around structurally significant areas rather than diffuse uniformly 
across the volume. This design keeps the original efficiency advantage of explicit 
Gaussian primitives, while changing the densification signal from a local gradient 
heuristic to a globally attenuation-aware sampling process. As shown in 
\fref{fig:teaser}, the resulting reconstruction better preserves sharp structural 
boundaries and fine anatomical details under sparse-view acquisition.

\paragraph{Semantic regularization.}
Although density-guided MCMC improves structural allocation under sparse-view 
conditions, the reconstruction remains underconstrained because the available 
projections still provide limited geometric evidence. We therefore incorporate 
semantic regularization based on zero-shot features from Segment Anything in 
Medical Images~\cite{MedSAM}. The motivation is that foundation-model features 
capture higher-level anatomical cues, such as boundary continuity and structural 
connectivity, which are less sensitive to local intensity fluctuations than raw 
pixel losses. By enforcing cross-view consistency in this semantic feature space, 
the regularization introduces additional structural constraints beyond 
projection-based supervision, leading to more reliable recovery of dental morphology.

\noindent\textbf{Contribution.}
We propose AD-GS, an anatomical density-guided framework for dental/maxillofacial sparse-view CBCT reconstruction. First, we reformulate Gaussian densification as an attenuation-aware MCMC sampling process, allowing Gaussian primitives to be adaptively allocated to high-attenuation dental hard tissues (enamel, cortical bone) rather than relying only on unstable sparse-view gradients. Second, we introduce the \emph{SSR (Semantic Structure Regularization)} module based on zero-shot MedSAM features to enforce multi-view consistency and constrain reconstruction under limited projections. Third, we validate the effectiveness of AD-GS on the ToothFairy3 dental CBCT benchmark through reconstruction quality, downstream tooth segmentation, robustness analysis, and ablation studies, demonstrating improved structural fidelity and boundary preservation under sparse-view settings within the dental CBCT setting.

\section{Related Work}
\label{sec:related_work}

\noindent\textbf{Sparse-View CBCT Reconstruction.} 
Sparse-view reconstruction aims to recover a high-quality 3D volume from a limited number of X-ray projections to minimize radiation dose \cite{ALARA}. Classical analytical methods, such as the Feldkamp-Davis-Kress (\fdk) algorithm, are computationally efficient but suffer from severe streaking artifacts and noise under sparse sampling \cite{R2}. Iterative reconstruction techniques, such as \sart and \asdpocs, incorporate hand-crafted regularizers but often show limited performance in sparse-view settings, as these priors may not fully compensate for the missing projection data. Recently, deep learning-based methods \cite{Difgs} have demonstrated impressive performance by learning mappings from sinograms to volumes. However, these supervised approaches require large-scale paired datasets, which are difficult to acquire in clinical practice, and often struggle with domain shifts across different imaging protocols or anatomical pathologies.

\noindent\textbf{Neural Representations and 3DGS in Tomography.}
Neural Radiance Fields (NeRF) have inspired a new paradigm of coordinate-based neural representations for CT reconstruction, such as \naf and \saxnerf. These methods optimize a continuous attenuation field through differentiable volumetric rendering, reducing the reliance on massive training data. However, the high computational overhead of dense ray sampling in NeRF limits its clinical utility. To address this, 3D Gaussian Splatting (3DGS) has emerged as an efficient explicit representation. Recent works like \rrgs and X-Field \cite{wangx} adapt 3DGS to CT by incorporating the Beer-Lambert law \cite{beer1,beer2}. Despite their efficiency, these methods rely on heuristic gradient-driven densification. In sparse-view settings, the rendering gradients often become noisy and inconsistent across views, leading to "gradient collision" \cite{AbsGS}. This causes unstable primitive allocation and fails to recover sharp boundaries of high-density dental structures.

\noindent\textbf{Probabilistic Sampling and Semantic Guidance.}
To move beyond heuristic densification, recent research has explored reformulating Gaussian management as a probabilistic process. Specifically, 3DGS-MCMC \cite{MCMC} interprets the Gaussian set as samples from a posterior distribution, using a Markov Chain Monte Carlo birth-death process to ensure stable scene coverage. While effective for surface-based RGB reconstruction, standard MCMC does not account for the volumetric attenuation characteristics of CBCT. Our work extends this by introducing an anatomical density-guided sampling strategy, biasing the MCMC process toward high-attenuation regions. 

Furthermore, foundation models like Segment Anything in Medical Images (MedSAM) \cite{MedSAM} have shown remarkable zero-shot capabilities in extracting high-level anatomical features. Recent studies in 3D reconstruction have begun to leverage such semantic features to enforce multi-view consistency. Unlike existing methods that use semantics for segmentation alone, AD-GS incorporates MedSAM-derived zero-shot features as a structural regularization term. This provides an additional geometric constraint that helps resolve the inherent ambiguity of sparse-view reconstruction, ensuring topological continuity even when projection information is extremely limited.

\begin{figure}[t]
    \centering
    \includegraphics[width=1.0\textwidth]{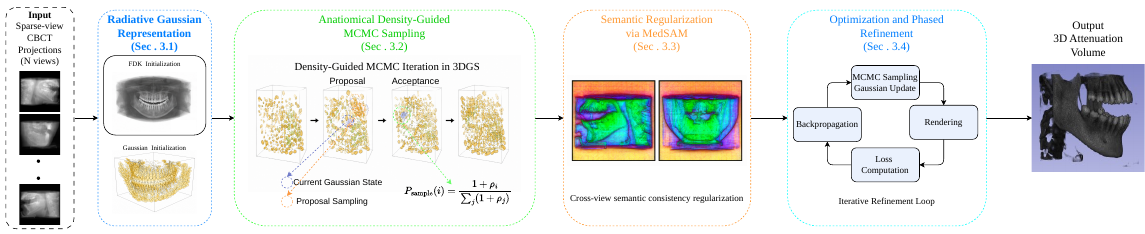} 
   
    \caption{Overview of the AD-GS framework. Starting from FDK initialization, Radiative Gaussians are optimized via projection-consistency. The reconstruction is enhanced by anatomical density-guided MCMC sampling for dental structures and MedSAM-based semantic regularization for structural connectivity.}
    \label{fig:pipeline}
\end{figure}

\section{Method}

As illustrated in Fig.~\ref{fig:pipeline}, our framework reconstructs a 3D attenuation volume from sparse-view CBCT projections by representing the scene with \emph{Radiative Gaussians} (Sec.~\ref{sec:background}), initialized via \fdk. To accurately capture dental morphology, we propose \emph{Anatomical Density-Guided MCMC Sampling}  (Sec.~\ref{sec:mcmc_sampling}), which adaptively manages Gaussian proliferation to concentrate representational capacity on high-attenuation structures. To mitigate geometric ambiguities inherent in sparse views, we further integrate \emph{Semantic Regularization via MedSAM} (Sec.~\ref{sec:regularization}) to enforce structural consistency and connectivity. Finally, the entire pipeline is governed by a phased refinement schedule (Sec.~\ref{sec:optimization}) that orchestrates the transition from global geometry to local detail recovery.

\subsection{Radiative Gaussians}
\label{sec:background}

In \rrgs, the 3D volume is represented by $N$ radiative Gaussians. Each primitive contributes a local attenuation component, and the global attenuation field $\sigma(\mathbf{x})$ at position $\mathbf{x} \in \mathbb{R}^3$ is obtained by accumulating their contributions:
\begin{equation}
    \sigma(\mathbf{x}) = \sum_{i=1}^N \rho_i \exp \left( -\frac{1}{2}(\mathbf{x} - \mathbf{p}_i)^\top \boldsymbol{\Sigma}_i^{-1} (\mathbf{x} - \mathbf{p}_i) \right),
\end{equation}
where $\rho_i, \mathbf{p}_i, \boldsymbol{\Sigma}_i$ are the peak density, centroid, and covariance matrix. The density parameter $\rho_i$ controls the attenuation strength of a primitive, while $\mathbf{p}_i$ determines its spatial location. The covariance matrix $\boldsymbol{\Sigma}_i$ controls the anisotropic spatial extent of the primitive, allowing the representation to model elongated or compact anatomical patterns. To ensure positive semi-definiteness during optimization, $\boldsymbol{\Sigma}_i$ is decomposed into rotation $\mathbf{R}_i$ and scaling $\mathbf{S}_i$ as $\boldsymbol{\Sigma}_i = \mathbf{R}_i \mathbf{S}_i \mathbf{S}_i^\top \mathbf{R}_i^\top$.

For a ray $\mathbf{r}(t) = \mathbf{o} + t\mathbf{d}$ with origin $\mathbf{o}$ and direction $\mathbf{d}$, the projection $I(\mathbf{r})$ is the line integral $I(\mathbf{r}) = \int \sigma(\mathbf{r}(t)) dt$. This formulation directly matches the tomographic imaging process, where the detector records accumulated attenuation along X-ray paths. Since the Gaussian representation is differentiable with respect to density, position, rotation, and scaling, reconstruction can be performed by minimizing the discrepancy between rendered projections and log-transformed X-ray images. In sparse-view CBCT, however, differentiability alone is insufficient: the allocation of primitives must also reflect where anatomical detail is most important. This motivates the density-guided sampling strategy introduced next.

\subsection{Anatomical Density-Guided MCMC Sampling} 
\label{sec:mcmc_sampling}

To resolve the limitations of heuristic densification in sparse-view CBCT, we reformulate Gaussian proliferation as a MCMC sampling process~\cite{MCMC}. In this framework, the Gaussian set $\mathbf{g}$ is treated as a sample from the distribution $\mathcal{P}(\mathbf{g}) \propto \exp(-\mathcal{E}(\mathbf{g}))$, where $\mathcal{E}(\mathbf{g})$ is the energy function representing the scene's negative log-likelihood. Instead of deterministically cloning or splitting primitives according to hand-designed gradient thresholds, the representation evolves through a birth-death process. ``Dead'' Gaussians with negligible contribution are stochastically relocated to ``live'' regions, allowing the model to maintain stable scene coverage while avoiding uncontrolled growth in primitive number.

For a candidate primitive, the relocation probability is:
\begin{equation}
    P(\text{spawn at } j) = \frac{w_j}{\sum_{k \in \text{live}} w_k},
\label{eq:sample_weight}
\end{equation}
where $w_j$ represents an importance score. Relocation is a probability-preserving state transition where parameters are adjusted to ensure $\mathcal{P}(\mathbf{g}^{\text{new}}) \approx \mathcal{P}(\mathbf{g}^{\text{old}})$. This probabilistic view is attractive for sparse-view reconstruction because it decouples primitive management from a single deterministic gradient threshold. However, the standard MCMC relocation strategy~\cite{MCMC} is tailored for surface-based RGB reconstruction, where sampling is concentrated on visible interfaces. CBCT reconstruction is different: the target is a volumetric attenuation field, and important structures are not necessarily defined by visible RGB surfaces, but by material density and anatomical attenuation transitions.

We observe a critical anatomical prior: high-density structures (e.g., enamel and cortical bone) exhibit sharp attenuation transitions while occupying a small volume. Under sparse-view conditions, allocating representational capacity to these regions is crucial for structural fidelity. To leverage this prior, we propose Anatomical Density-Guided Sampling. We define the sampling probability $\omega_i$ by normalizing the log-transformed density $\rho_i$ across all $N$ primitives:
\begin{equation}
    \omega_i = \text{Softmax}(\log(1 + \rho_i)) = \frac{1 + \rho_i}{\sum_{j=1}^N (1 + \rho_j)}.
\end{equation}
The logarithmic form avoids excessively sharp concentration on a few very dense primitives while still increasing the probability of sampling around high-attenuation regions. By incorporating $\omega_i$ into MCMC sampling, Gaussian primitives are encouraged to proliferate and refine in high-attenuation areas. This strategy resolves ``gradient collision''~\cite{AbsGS} by providing an additional density-aware allocation signal when sparse-view gradients become unreliable. Consequently, AD-GS preserves topological fidelity and supports intricate dental structure reconstruction, especially around thin enamel boundaries and interproximal spaces.

\begin{figure}[t]
    \centering
    \includegraphics[width=0.9\textwidth]{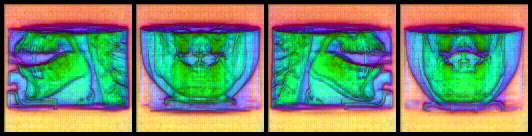} 
    \caption{Visualization of MedSAM features via PCA. The compressed semantic maps delineate clear dental structures and boundaries, serving as a structural prior beyond raw intensities.}
    \label{fig:sem}
\end{figure}

\subsection{Semantic Regularization via MedSAM Features}
\label{sec:regularization}

Although density-guided MCMC improves structural allocation, sparse-view CBCT remains underconstrained. \Ie, intensity supervision (e.g., $\mathcal{L}_1$ or SSIM) often fails at low-contrast boundaries, leading to geometric ambiguity and disconnected structures. This issue is particularly important for dental CBCT, where a visually small boundary error may change the topology of a tooth surface or close a narrow gap between adjacent teeth. Therefore, projection-level intensity consistency alone is insufficient and should be complemented with a constraint that is more sensitive to anatomical structure.

To address this, we incorporate the \emph{SSR} (Semantic Structure Regularization) module using MedSAM~\cite{MedSAM} features, which are invariant to intensity fluctuations. MedSAM is trained to encode medical image structures and thus provides feature responses that capture boundary organization and regional consistency beyond raw grayscale values. By enforcing consistency between rendered and ground-truth projections in this feature space, the regularization provides structural constraints beyond intensity supervision. Importantly, this does not require manual segmentation labels during reconstruction; the zero-shot feature encoder is used only to define a semantic alignment signal.

The MedSAM encoder extracts features $\mathbf{f}_{p, gt}$ and $\hat{\mathbf{f}}_p$ at pixel $p$, where $\hat{\mathbf{f}}_p, \mathbf{f}_{p, gt} \in \mathbb{R}^C(C = 352)$. 
By computing the PCA projection matrix $\mathbf{V}$ based solely on the distribution of ground truth features $\mathbf{f}_{p, gt}$, we obtain latent vectors $\hat{\mathbf{z}}_p = \mathbf{V} \hat{\mathbf{f}}_p$ and $\mathbf{z}_{p, gt} = \mathbf{V} \mathbf{f}_{p, gt}$, where $\mathbf{z} \in \mathbb{R}^{d}(d=16)$. As illustrated in \fref{fig:sem}, mapping the first three principal components to the RGB space reveals that these zero-shot features effectively capture intricate anatomical boundaries and structural hierarchies that are often ambiguous in raw intensity projections. This PCA compression not only reduces computational cost and suppresses feature-channel redundancy but also preserves the dominant semantic variations required for stable multi-view alignment. We then minimize the cosine distance:
\begin{equation}
    \mathcal{L}_{sem} = \frac{1}{HW} \sum_{p=1}^{HW} \left( 1 - \frac{\hat{\mathbf{z}}_p \cdot \mathbf{z}_{p, gt}}{\|\hat{\mathbf{z}}_p\|_2 \|\mathbf{z}_{p, gt}\|_2} \right).
\end{equation}
This loss encourages primitives to consolidate at anatomical boundaries, ensuring geometric consistency and connectivity under limited projections. Since the cosine distance focuses on feature direction rather than absolute magnitude, it is less sensitive to global intensity scaling and better suited to enforcing structural agreement between rendered and measured projections.

\subsection{Optimization and Phased Refinement}
\label{sec:optimization}

We optimize attributes $\Theta = \{x, s, r, \rho\}$ (position, scaling, rotation, density). The total objective is:
\begin{equation}
    \mathcal{L}_{total} = \mathcal{L}_{img} + \lambda_{tv} \mathcal{L}_{tv} + \lambda_{iso} \mathcal{L}_{iso} + \lambda_{sem} \mathcal{L}_{sem},
\end{equation}
Where, $\mathcal{L}_{\text{img}}$ combines $\mathcal{L}_1$ and SSIM to ensure pixel-level fidelity, while $\mathcal{L}_{\text{tv}}$ suppresses noise to encourage piecewise smoothness in the recovered attenuation field. To maintain physically plausible primitive shapes, $\mathcal{L}_{\text{iso}}$~\cite{rank} acts as an isotropic regularization that penalizes excessive scaling ratios, effectively preventing the formation of ``needle-like'' artifacts. Such artifacts are common in sparse-view reconstruction because elongated primitives can explain a small number of projections while failing to represent a plausible 3D structure. Finally, $\mathcal{L}_{\text{sem}}$ (detailed in Sec.~\ref{sec:regularization}) provides high-level anatomical guidance.

We employ a coarse-to-fine optimization schedule to ensure stability. In the early stage, the geometric backbone is established using $\mathcal{L}_{\text{img}}$, $\mathcal{L}_{\text{tv}}$, and $\mathcal{L}_{\text{iso}}$ to recover global structures and suppress severe sparse-view artifacts. During this stage, AD-GS focuses on obtaining a reliable attenuation distribution from which density-guided sampling can identify meaningful high-attenuation regions. Once the volume becomes stable, $\mathcal{L}_{\text{sem}}$ is activated to refine structural boundaries and connectivity. This phased refinement prevents semantic features from forcing premature local alignment before the coarse geometry is reliable, while still allowing MedSAM-based guidance to improve high-frequency anatomical details in the later stage.

\begin{figure}[t]
    \centering
    \includegraphics[width=1.0\textwidth]{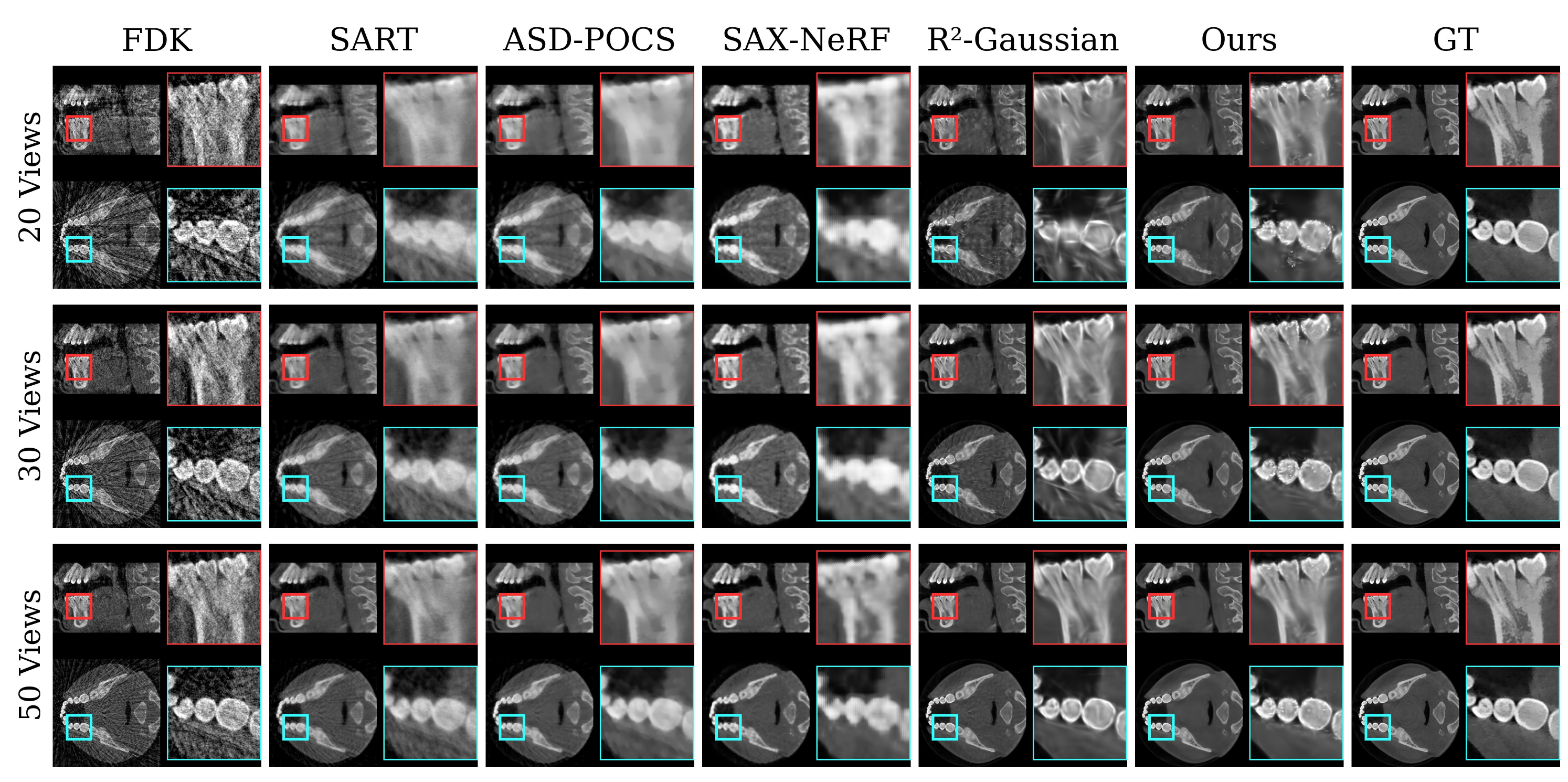} 
    \caption{Comparison on CBCT reconstruction under different numbers of views. Ours are closest to the GT, especially near the region containing hard tissue.}
    \label{fig:volume_result}
\end{figure}

\begin{table}[t]
    \centering
    \caption{Quantitative comparison of tooth reconstruction performance across different view counts. The best results are highlighted in \textbf{bold}, and the second-best results are \underline{underlined}. DSC, NSD, and clDice are reported in (\%), while ASSD is in ($10^{-2}$mm) and PSNR is in dB.}
    \label{tab:comprehensive_results_transposed}
    \footnotesize
    \resizebox{\textwidth}{!}{
        \setlength{\tabcolsep}{6pt}
        \begin{tabular}{cl cccc cc}
            \toprule
            \textbf{Views} & \textbf{Metric} & \textbf{\fdk} & \textbf{\sart} & \textbf{\asdpocs} & \textbf{\saxnerf} & \textbf{\rrgs} & \textbf{Ours} \\
            \midrule
            
            \multirow{5}{*}{\textbf{20}} 
            & DSC $\uparrow$      & \underline{80.39} & 76.73 & 78.61 & 25.25 & 78.75 & \textbf{88.26} \\
            & NSD $\uparrow$      & \underline{83.17} & 79.23 & 82.09 & 30.27 & 82.17 & \textbf{93.69} \\
            & clDice $\uparrow$      & \underline{79.94} & 76.98 & 79.50 & 28.13 & 78.48 & \textbf{90.05} \\
            & ASSD $\downarrow$ & \underline{88.50} & 94.65 & 94.06 & 392.47 & 91.14 & \textbf{44.03} \\
            & PSNR $\uparrow$ & 17.67 & 27.67 & 27.44 & 27.19 & \underline{28.81} & \textbf{30.78} \\

            \midrule
            
            \multirow{5}{*}{\textbf{30}} 
            & DSC $\uparrow$      & 88.02 & 86.01 & 87.59 & 62.43 & \underline{89.97} & \textbf{92.89} \\
            & NSD $\uparrow$      & 91.22 & 89.47 & 91.56 & 66.32 & \underline{94.49} & \textbf{98.01} \\
            & clDice $\uparrow$      & 88.52 & 86.14 & 86.43 & 62.18 & \underline{91.81} & \textbf{95.79} \\
            & ASSD $\downarrow$ & 49.52 & 58.16 & 45.79 & 171.10 & \underline{32.68} & \textbf{20.70} \\
            & PSNR $\uparrow$ & 19.61 & 29.19 & 28.85 & 28.64 & \underline{31.60} & \textbf{33.11} \\

            \midrule
            
            \multirow{5}{*}{\textbf{50}} 
            & DSC $\uparrow$      & 92.56 & 90.88 & 92.09 & 78.97 & \underline{93.87} & \textbf{94.33} \\
            & NSD $\uparrow$      & 95.86 & 94.46 & 95.99 & 83.44 & \underline{97.88} & \textbf{98.57} \\
            & clDice $\uparrow$      & 92.31 & 90.96 & 92.37 & 78.13 & \underline{95.85} & \textbf{96.19} \\
            & ASSD $\downarrow$ & 26.27 & 32.16 & 25.42 & 93.90 & \underline{18.61} & \textbf{16.93} \\
            & PSNR $\uparrow$ & 21.80 & 31.54 & 30.91 & 30.05 & \underline{34.22} & \textbf{34.38} \\

            \bottomrule
        \end{tabular}
    }
\end{table} 
\begin{figure}[t]
    \centering
    \includegraphics[width=0.9\textwidth]{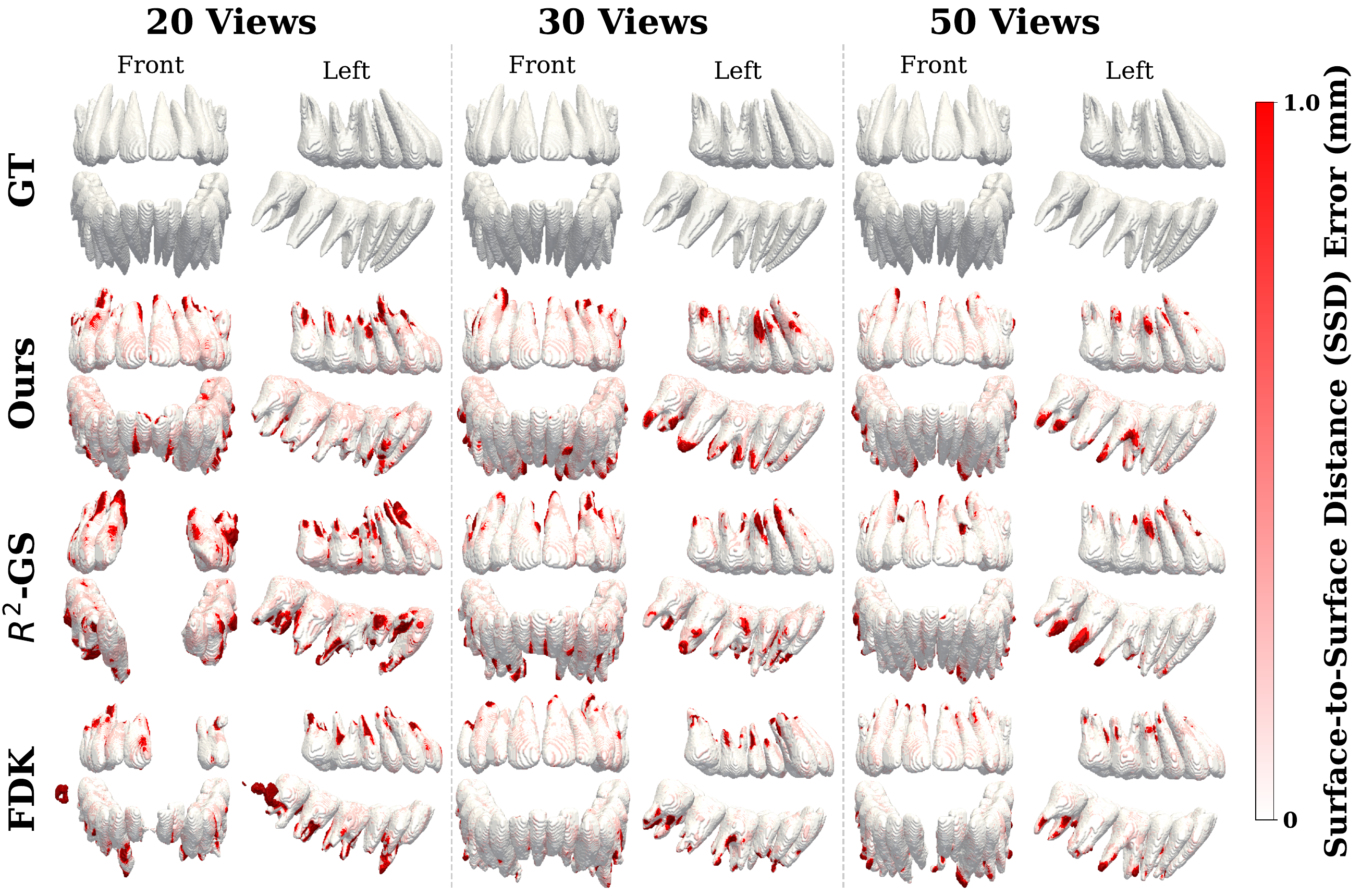} 
    \caption{Evaluation on Downstream Task: Tooth Segmentation~\cite{toothseg}. Side and frontal views of segmentation results obtained from different CBCT reconstructions. Geometric errors with respect to ground truth (GT), measured by SSD, are visualized using color maps. Our volume yields the most complete tooth segmentation and lowest SSD.
    }
    \label{fig:seg_result}
\end{figure}

\begin{table}[t]
    \centering
    \caption{Quantitative evaluation under different initializations, noise levels, and ablation settings.}
    \label{tab:excel_style_results}
        \footnotesize
        \resizebox{\textwidth}{!}{
            \setlength{\tabcolsep}{3pt}
            \begin{tabular}{ll cccc cccc cccc} 
                \toprule
                \multicolumn{2}{l}{Views} & \multicolumn{4}{c}{\textbf{20}} & \multicolumn{4}{c}{\textbf{30}} & \multicolumn{4}{c}{\textbf{50}} \\
                \cmidrule(lr){3-6} \cmidrule(lr){7-10} \cmidrule(lr){11-14}
                
                \multicolumn{2}{l}{Metric} 
                & \scalebox{0.85}{DSC$\uparrow$} & \scalebox{0.85}{NSD$\uparrow$} & \scalebox{0.85}{clD$\uparrow$} & \scalebox{0.85}{ASSD$\downarrow$} 
                & \scalebox{0.85}{DSC$\uparrow$} & \scalebox{0.85}{NSD$\uparrow$} & \scalebox{0.85}{clD$\uparrow$} & \scalebox{0.85}{ASSD$\downarrow$} 
                & \scalebox{0.85}{DSC$\uparrow$} & \scalebox{0.85}{NSD$\uparrow$} & \scalebox{0.85}{clDice$\uparrow$} & \scalebox{0.85}{ASSD$\downarrow$} \\
                
                \midrule
                \multicolumn{14}{l}{\textit{Exp 1: Robustness against Initialization (Tested on Noisy Data)}} \\
                \midrule
                \multirow{2}{*}{\rrgs} 
                & FDK    & 80.70 & 83.50 & 78.60 & 74.14 & 91.61 & 96.02 & 93.68 & 26.08 & 95.57 & 99.11 & 97.52 & 12.87 \\
                & Random & 76.27 & 77.73 & 74.98 & 98.29 & 90.83 & 95.08 & 92.79 & 29.10 & 95.36 & 99.08 & 97.45 & 13.24 \\
                \addlinespace[2pt]
                \multirow{2}{*}{\textbf{Ours}} 
                & FDK    & \textbf{90.53} & \textbf{95.30} & \textbf{92.00} & \textbf{28.75} & \underline{94.13} & \textbf{98.83} & \textbf{96.42} & \underline{16.89} & \underline{95.74} & \underline{99.56} & \textbf{97.79} & \underline{12.45} \\
                & Random & \underline{89.98} & \underline{94.55} & \underline{91.50} & \underline{29.68} & \textbf{94.27} & \underline{98.67} & \underline{96.20} & \textbf{16.69} & \textbf{95.92} & \textbf{99.66} & \textbf{97.79} & \textbf{11.97} \\
                
                \midrule
                \multicolumn{14}{l}{\textit{Exp 2: Robustness against Noise (Tested with FDK Initialization)}} \\
                \midrule
                \multirow{2}{*}{\rrgs} 
                & Clean & 88.00 & 92.01 & 88.47 & 39.17 & 93.32 & 97.29 & 95.18 & 19.87 & 95.47 & 99.07 & 97.18 & 12.83 \\
                & Noisy & 80.70 & 83.50 & 78.60 & 74.14 & 91.61 & 96.02 & 93.68 & 26.08 & 95.57 & 99.11 & 97.52 & 12.87 \\
                \addlinespace[2pt]
                \multirow{2}{*}{\textbf{Ours}} 
                & Clean & \textbf{92.10} & \textbf{96.66} & \textbf{93.86} & \textbf{23.88} & \textbf{94.96} & \textbf{99.10} & \textbf{97.20} & \textbf{14.56} & \textbf{96.22} & \textbf{99.71} & \textbf{98.03} & \textbf{11.15} \\
                & Noisy & \underline{90.53} & \underline{95.30} & \underline{92.00} & \underline{28.75} & \underline{94.13} & \underline{98.83} & \underline{96.42} & \underline{16.89} & \underline{95.74} & \underline{99.56} & \underline{97.79} & \underline{12.45} \\

                \midrule
                \multicolumn{14}{l}{\textit{Exp 3: Ablation Study (Tested with Noisy Data + FDK Initialization)}} \\
                \midrule
                \multirow{3}{*}{\textbf{Ours}} 
                & Grad-driven & 89.83 & 92.79 & 88.79 & 34.10 & 93.89 & 98.04 & 95.00 & 18.01 & \textbf{95.81} & \underline{99.53} & \textbf{98.10} & \textbf{12.12} \\
                & w/o SSR     & \underline{90.15} & \underline{93.33} & \underline{91.42} & \underline{32.24} & \underline{93.97} & \underline{98.32} & \underline{95.23} & \underline{17.44} & 95.71 & 99.42 & 97.29 & \underline{12.25} \\
                & Full model  & \textbf{90.53} & \textbf{95.30} & \textbf{92.00} & \textbf{28.75} & \textbf{94.13} & \textbf{98.83} & \textbf{96.42} & \textbf{16.89} & \underline{95.74} & \textbf{99.56} & \underline{97.79} & 12.45 \\
                \bottomrule
            \end{tabular}
        }
\end{table}

\section{Experiment}

\subsection{Experimental Settings}

\noindent\textbf{Dataset.} We randomly sample 10 cases from the 37 available CBCT scans in the ToothFairy3 dataset~\cite{bolelli2025segmenting,bolelli2024toothfairy,lumetti2024enhancing}. This dataset contains dental and maxillofacial structures with complex hard-tissue geometry, making it suitable for evaluating whether sparse-view reconstruction preserves clinically relevant anatomy. Following \rrgs, we simulate $20$, $30$, and $50$ projection views using \textit{TIGRE} \cite{biguri2016tigre} along a circular trajectory with uniformly spaced angles. These view counts cover challenging sparse settings and a relatively denser setting, allowing us to analyze how different methods behave as projection constraints become gradually more sufficient. To better approximate real-world clinical conditions, we further inject Poisson noise ($10^5$ photons) and Gaussian noise ($\sigma = 0.01$) into the simulated projections.

\noindent\textbf{Implementation Details.} \ours is trained on an NVIDIA RTX 4090 GPU using the Adam optimizer. Each scene takes approximately 30 minutes to converge. The hyperparameters are set as $\lambda_{\text{ssim}} = 0.25$, $\lambda_{\text{tv}} = 0.05$, $\lambda_{\text{sem}} = 0.1$, and $\lambda_{\text{iso}} = 0.01$. We keep the same hyperparameter setting across view counts to evaluate the robustness of the proposed formulation rather than tuning the method for each sparsity level. During optimization, projection consistency provides the primary reconstruction signal, while density-guided MCMC and semantic regularization are responsible for improving allocation and structural coherence. This setting reflects the intended use case of AD-GS: stable sparse-view reconstruction without relying on case-specific manual adjustment.

\subsection{Comparison with Existing Methods}

\noindent\textbf{Baselines.} We compare our method against traditional reconstruction algorithms, including \fdk, \sart, and \asdpocs, as well as the NeRF-based method \saxnerf and the 3DGS-based method \rrgs. These baselines cover complementary reconstruction paradigms: analytical reconstruction, iterative optimization, neural implicit reconstruction, and explicit Gaussian reconstruction. This comparison allows us to separate the benefits of the 3DGS representation from the benefits introduced by anatomical density-guided sampling and semantic regularization.

\noindent\textbf{Evaluation Metrics.} Following \rrgs and \saxnerf, we adopt {PSNR} to evaluate overall reconstruction fidelity. PSNR measures global intensity agreement and is useful for checking whether the reconstructed attenuation field matches the reference volume on average. However, \emph{CBCT is primarily used for clinical analysis rather than visual inspection.} A reconstruction with high global smoothness may still miss thin structures or blur dental boundaries, which can be problematic for diagnosis and treatment planning. Therefore, to assess practical utility under sparse scanning, we further employ \toothseg on reconstructed CBCT volumes and report four segmentation-based metrics: {DSC} (\%), {NSD} (\%, 1.0 mm tolerance), {clDice} (\%), and {ASSD} ($10^{-2}$ mm). These metrics measure volumetric overlap, boundary agreement, topological connectivity, and surface distance, respectively, providing a more anatomy-aware evaluation of reconstruction quality.

As shown in \Tref{tab:comprehensive_results_transposed}, we quantitatively compare existing methods with \ours. Under relatively dense sampling (50 views), all methods achieve comparable performance, indicating that the reconstruction task becomes less ambiguous when sufficient angular information is available. However, when the number of projections is reduced to 30 or 20, traditional reconstruction algorithms exhibit a noticeable degradation, which is also reflected in the qualitative results (see Fig.~\ref{fig:volume_result}). Specifically, blurred and overly smoothed structures appear in high-density anatomical regions, as highlighted in the zoomed-in views. This degradation mainly arises because classical methods such as FDK suffer from severe ambiguity under sparse-view conditions, where the inverse problem admits multiple feasible solutions.

Compared with the SOTA methods \rrgs, our method achieves higher PSNR values across different view counts. This improvement suggests that density-guided sampling does not simply sharpen local boundaries at the expense of global reconstruction fidelity. Instead, it improves the allocation of representational capacity while maintaining projection consistency. The qualitative results in Fig.~\ref{fig:volume_result} further support this observation: AD-GS preserves noticeably sharper structures in clinically relevant bone regions, particularly under sparse-view settings. In contrast, competing methods either amplify artifacts or smooth high-attenuation boundaries, making fine dental morphology less distinguishable.

It is also important to interpret PSNR together with anatomy-aware metrics. A reconstruction may obtain a reasonable PSNR by recovering low-frequency attenuation patterns while still failing to preserve thin boundaries or small gaps. For CBCT, such local errors can be more clinically relevant than a small global intensity difference. Therefore, the segmentation-based metrics in \Tref{tab:comprehensive_results_transposed} provide complementary evidence. AD-GS consistently improves DSC, NSD, clDice, and ASSD, showing that the reconstructed volumes are more suitable for downstream structural analysis.

We further evaluate downstream clinical performance (tooth segmentation~\cite{toothseg}) on CBCT volumes reconstructed by different methods, as shown in \fref{fig:seg_result}. To quantify structural accuracy, we visualize the surface-to-surface distance (SSD) between the segmented tooth surfaces and the ground truth. Our method produces the most complete tooth segmentation and achieves the lowest SSD. The error maps show fewer regions with large over-estimation or under-estimation, indicating that AD-GS better preserves tooth shape and boundary continuity. These findings are consistent with the quantitative results reported in \Tref{tab:comprehensive_results_transposed}, where \ours attains the smallest average SSD (ASSD) and outperforms existing methods across the other three segmentation metrics. These results demonstrate that our ADS effectively preserves clinically critical anatomical structures under sparse-view.

\subsection{Robustness against Initialization \& Noise}

\Tref{tab:excel_style_results} evaluates initialization sensitivity for \rrgs and \ours under 20 views. With random point cloud instead of \fdk reconstruction, \rrgs DSC drops from $80.70\%$ to $76.27\%$, while \ours maintains $89.98\%$. This indicates that \rrgs depends on structured initialization for reliable gradients. When the initialization is less informative, gradient-driven clone-and-split may propagate Gaussians toward suboptimal regions because the sparse-view supervision cannot consistently identify where primitives should be refined. By contrast, ADS allocates Gaussians to high-density regions from the outset, reducing reliance on gradient refinement and improving robustness.

Exp.2 further shows that \ours outperforms \rrgs under noisy projections across view counts, highlighting noise robustness. This behavior is consistent with the design of AD-GS. Noise perturbs projection intensities and can make local gradients less reliable, but the anatomical density-guided sampling mechanism provides an additional prior that remains tied to the recovered attenuation structure. Meanwhile, semantic regularization encourages feature-level structural agreement rather than only pixel-level matching. Together, these two components make the optimization less sensitive to noisy projection signals and more stable under realistic sparse-view acquisition.

\subsection{Ablation study}

As shown in the bottom rows of \Tref{tab:excel_style_results}, we validate the effectiveness of the ADS and SSR modules. Replacing ADS with vanilla gradient-driven densification results in consistent performance degradation across all metrics, highlighting the necessity of anatomically guided Gaussian allocation. This confirms that density-aware initialization plays a critical role in stabilizing reconstruction under sparse-view conditions. The performance drop is especially meaningful because the baseline still uses the same overall representation and projection supervision; therefore, the difference can be attributed to how primitives are allocated during optimization.

In addition, removing the SSR module primarily affects topological
integrity, and this effect is most pronounced under the hardest 20-view
noisy setting, where NSD drops from $95.30$ to $93.33$ and ASSD
increases from $28.75$ to $32.24$ ($10^{-2}$mm). This concentration
under extreme sparsity indicates that semantic regularization is most
beneficial precisely when projection constraints alone cannot resolve
boundary connectivity, which is the regime SSR is designed to act in.
As the number of views increases to 30 and 50, the gap between the full
model and the w/o-SSR variant narrows because projection data become
more sufficient and the reconstruction problem becomes less ill-posed.
This trend is consistent with the intended role of the SSR module:
semantic priors are not a replacement for projection evidence, but a
complementary structural constraint whose marginal contribution
naturally saturates as the angular coverage increases.


\begin{table}[t]
\centering
\caption{Training time (seconds) on an NVIDIA RTX 4090. AD-GS adds manageable overhead over R$^2$-Gaussian while remaining an order of magnitude faster than NeRF-based methods.}
\label{tab:training_time}
\footnotesize              
\setlength{\tabcolsep}{3pt} 
\begin{tabular}{@{}cl|rrr@{}} 
\toprule
\textbf{Cat.} & \textbf{Method} & \textbf{20 Views} & \textbf{30 Views} & \textbf{50 Views} \\
\midrule
\multirow{3}{*}{Trad.} 
    & \fdk & 1.3 & 1.7 & 1.2 \\
    & \sart & 77.5 & 115.1 & 195.0 \\
    & \asdpocs & 79.7 & 122.0 & 198.0 \\
\midrule
NeRF-based  & \saxnerf & 11,210 & 16,092 & 25,828 \\
\midrule
\multirow{3}{*}{GS-based} 
    & \rrgs & 854 & 855 & 833 \\
    & AD-GS (w/o SSR) & 1,439 & 1,413 & 1,348 \\
    & \textbf{AD-GS (Full)} & 1,650 & 1,633 & 1,607 \\
\bottomrule
\end{tabular}
\end{table} 

\subsection{Discussion}
\label{sec:discussion}

\noindent\textbf{Mechanism of Density-Guided Allocation.} 
The core advantage of AD-GS lies in its ability to decouple Gaussian proliferation from unstable sparse-view gradients. As shown in Fig.~\ref{fig:volume_result}, while traditional gradient-driven methods (\eg, \rrgs) suffer from "gradient collision" where inconsistent projections blur high-frequency boundaries, our ADS strategy uses the recovered attenuation density as a structural prior. By preferentially allocating representational capacity to high-density dental tissues (enamel and cortical bone), AD-GS maintains sharp structural boundaries even when the number of views is reduced to 20, effectively resolving the geometric ambiguity inherent in sparse-view acquisition.

\noindent\textbf{Structural Consistency via Semantic Priors.} 
The SSR module provides a higher-level constraint that intensity-based losses (\eg, $L_1$) lack. In sparse settings, different volumetric configurations may yield similar projection residuals, leading to disconnected or merged tooth structures. By enforcing multi-view consistency in the MedSAM feature space, AD-GS ensures that the reconstructed volume adheres to anatomical connectivity. This is reflected in the significant gains in \textit{clDice} and \textit{ASSD} (\Tref{tab:comprehensive_results_transposed}), which measure topological integrity rather than just pixel-level similarity. The ablation study confirms that this semantic guidance is most critical at extreme sparsity (20 views) and naturally transitions to projection-driven refinement as more views become available.

\noindent\textbf{Computational Efficiency and Practicality.} 
As quantified in \Tref{tab:training_time}, AD-GS strikes a favorable balance between reconstruction fidelity and computational cost. Compared to the NeRF-based SAX-NeRF, which requires over 7 hours for a 50-view reconstruction, AD-GS converges in approximately 27 minutes—a speedup of over 15$\times$. While our method introduces additional overhead compared to the baseline \rrgs due to MCMC sampling and MedSAM feature extraction, the training time remains relatively constant across different view counts. This efficiency is particularly advantageous in clinical dental workflows, where rapid 3D visualization is essential for chairside diagnostics and treatment planning. Ultimately, harmonizing superior reconstruction quality with high computational efficiency remains the ideal objective for clinical practice, ensuring that advanced algorithms are both high-performing and practically accessible.

\noindent\textbf{Clinical Utility beyond PSNR.} 
Our experiments highlight a gap between global fidelity (PSNR) and clinical utility. While a smoothed reconstruction might achieve a reasonable PSNR by matching average intensities, it may fail in downstream tasks like tooth segmentation. The superior performance of AD-GS in segmentation metrics (DSC $>$ 90\% at 20 views) demonstrates that preserving sharp boundaries and interproximal spaces is more vital for dental diagnostics than global smoothness. This robustness to noise and initialization (\Tref{tab:excel_style_results}) further underscores the potential of AD-GS for low-dose clinical workflows where acquisition conditions are often non-ideal.

\section{Conclusion}

We presented AD-GS, an anatomical density-guided 3DGS framework for
dental and maxillofacial sparse-view CBCT reconstruction. By
reformulating Gaussian densification as an attenuation-aware MCMC sampling
process, AD-GS adaptively allocates representational capacity to
high-attenuation dental hard tissues instead of relying solely on
unstable sparse-view gradients. Additionally, the incorporation of the
\emph{SSR module} provides critical structural constraints that ensure
multi-view consistency under limited projections. These two components
address complementary aspects of the sparse-view problem: ADS improves
where primitives are placed, while the SSR module improves how
reconstructed structures remain coherent across views. Experiments
demonstrate that our method achieves superior fidelity, sharper
anatomical boundaries, stronger downstream segmentation performance, and
enhanced robustness against initialization and noise compared to
state-of-the-art approaches on the ToothFairy3 dental CBCT benchmark.

\noindent\textbf{Limitation.} 
Our evaluation focuses on dental and maxillofacial CBCT, where the
target anatomy is dominated by high-attenuation hard tissues, and though
\ours achieves state-of-the-art performance over existing methods, it
still struggles under extremely sparse-view sampling conditions (\eg,
fewer than 10 projections), where the reconstruction problem becomes
severely ill-posed. In such cases, density-guided sampling and semantic
regularization alone are insufficient to reliably constrain the solution
space because too little projection evidence is available to
disambiguate fine 3D geometry. In addition, the use of a pre-trained
MedSAM encoder introduces extra feature-extraction overhead during
optimization. Future work will explore stronger physical priors, more
efficient semantic feature extraction, expanding evaluation to a larger
set of dental cases and datasets, and broader anatomical regions where
low-contrast soft tissues are also clinically important.

\section*{Acknowledgments}
{\sloppy
This work was supported in part by the Beijing Natural Science Foundation Project No.~L242025; in part by the Beijing Major Science and Technology Project under Contract No.~Z251100007125021, the Hebei Natural Science Foundation Project No.~242Q0101Z, and the Beijing-Tianjin-Hebei Basic Research Funding Program No.~F2024502017; and in part by the National Natural Science Foundation of China (Grant No.~62472044, U24B20155, U23B2052).\par}


%
%
%
%

\bibliography{ref}
\end{document}